\documentclass[journal]{IEEEtran}
\makeatletter
\def\endthebibliography{%
  \def\@noitemerr{\@latex@warning{Empty `thebibliography' environment}}%
  \endlist
}
\makeatother

\usepackage{cite}
\usepackage{amsmath}
\usepackage{threeparttable}
\usepackage{supertabular}
\usepackage{booktabs}
\usepackage{multirow}
\usepackage[T1]{fontenc}
\usepackage{float}
\usepackage{color}
\usepackage{amssymb}
\usepackage{makecell}
\ifCLASSINFOpdf
  \usepackage[pdftex]{graphicx}
  \usepackage{subfigure}
\else
\fi
\usepackage{amsmath}
\usepackage{bm}

\begin{document}
 
%
\title{A Weighted Heterogeneous Graph Based\\ Dialogue System}
%
%
%
\author{Xinyan Zhao,  Liangwei Chen, Huanhuan Chen,~\IEEEmembership{Senior Member,~IEEE} \thanks{X. Zhao and L. Chen are with the School of Data Science, University of Science and Technology of China, Hefei 230027, China (e-mail: sa516458@ustc.mail.edu.cn, CLW@ustc.mail.edu.cn). H. Chen is with the School of Computer Science and Technology, University of Science and Technology of China, Hefei 230027, China (e-mail:hchen@ustc.edu.cn)}
\thanks{Under review}}
\maketitle

\begin{abstract}

Knowledge based dialogue systems have attracted increasing research interest in diverse applications. However, for disease diagnosis, the widely used knowledge graph is hard to represent the symptom-symptom relations and symptom-disease relations since the edges of traditional knowledge graph are unweighted. Most research on disease diagnosis dialogue systems highly rely on data-driven methods and statistical features, lacking profound comprehension of symptom-disease relations and symptom-symptom relations. To tackle this issue, this work presents a weighted heterogeneous graph based dialogue system for disease diagnosis. Specifically, we build a weighted heterogeneous graph based on symptom co-occurrence and a proposed symptom frequency-inverse disease frequency. Then this work proposes a graph based deep Q-network (Graph-DQN) for dialogue management. By combining Graph Convolutional Network (GCN) with DQN to learn the embeddings of diseases and symptoms from both the structural and attribute information in the weighted heterogeneous graph, Graph-DQN could capture the symptom-disease relations and symptom-symptom relations better. Experimental results show that the proposed dialogue system rivals the state-of-the-art models. More importantly, the proposed dialogue system can complete the task with less dialogue turns and possess a better distinguishing capability on diseases with similar symptoms.

\end{abstract}


%
\IEEEpeerreviewmaketitle

\section{Introduction}

\IEEEPARstart{K}{nowledge} based dialogue systems have attracted increasing research interest in diverse applications \cite{EricKCM17, GhazvininejadBC18, zhao2020condition}. In the medical domain, dialogue systems could be used to find symptoms and make a diagnosis by conversing with patients (see Table \ref{demo} for an example). This type of dialogue systems have substantial potential to improve the efficiency of collecting information from patients \cite{KaoTC18} and assist general practitioners in clinical diagnosis.

As an emerging technology, knowledge graph is widely used in knowledge based dialogue systems\cite{ZhouYHZXZ18, MoonSKS19}. However, the edges of traditional knowledge graph is unweighted, which cannot represent the strength of the correlation between two nodes. Therefore, the form of knowledge graph may be not optimal for representing the symptom-disease relations and symptom-symptom relations, which play a key role in disease diagnosis. Most research on disease diagnosis dialogue systems make decisions based on the dialogue state without medical knowledge or highly rely on statistical features (i.e., conditional probabilities from symptoms to diseases) \cite{KaoTC18, WeiLPTCHWD18, XuZGLTL19, XiaZSLH20}. However, these features are hard to provide profound comprehension of symptom-disease relations and symptom-symptom relations.

\begin{table}[!t]
\centering
\caption{An example of disease diagnosis dialogue systems. Underlined phrases are symptoms.}
\label{demo}
\begin{tabular}{l}
\hline
\textbf{Self-report} \\
\hline 
Lately, my child began to \underline{cough} and has a \underline{fever} of 39 degrees.\\
\hline
\textbf{Dialogue}\\
Doctor: Does your child have a \underline{rash}?\\
Patient: Yes!\\
Doctor: Is your child \underline{anorexic}?\\
Patient: Yes!\\
Doctor: Your child may have \emph{hand foot mouth disease}.\\
\hline
\end{tabular}
\end{table}

This paper proposes a weighted heterogeneous graph based dialogue system for disease diagnosis, which is inspired by the fact that the reasoning process in the human brain is almost based on the graph extracted from experiences \cite{zhou2018graph}.  Specifically, we build a weighted heterogeneous graph based on clinical dialogues, which contains diseases and symptoms as nodes. The edge between two symptoms is built by symptom co-occurrence information. Moreover, we propose a symptom frequency-inverse disease frequency to measure the importance of a symptom to the diagnosis of a disease. Then this paper presents a graph based deep Q-network (Graph-DQN) for dialogue management\footnote{Dialogue management is the central controller of a dialogue system, and it decides which action to take in response to the user.}. Graph-DQN combines Graph Convolutional Network (GCN) \cite{KipfW17} with deep Q-network framework to learn the embedding of a disease or symptom in the constructed weighted heterogeneous graph. By integrating information about neighborhoods and edges of a disease or symptom node, the learnt embeddings could better capture symptom-disease relations and symptom-symptom relations.

Two public disease diagnosis dialogue datasets are used to evaluate the proposed dialogue system. Experimental results demonstrate that the proposed dialogue system rivals the existing state-of-the-art methods by modelling the constructed graph. Besides, our method can complete the diagnosis task with less dialogue turns and possesses better distinguishing capability on diseases with similar symptoms, which means a better user experience and higher competitiveness in real-world applications.

In summary, this paper makes the following contributions:
\begin{itemize}
\item This work depicts disease-symptom relations and symptom-symptom relations via a weighted heterogeneous graph that utilizes the symptom co-occurrence information and the proposed symptom frequency-inverse disease frequency.
\item This work proposes a weighted heterogeneous graph based dialogue system, in which Graph-DQN is proposed for the dialogue management.
\item Experimental analyses demonstrate that the proposed dialogue system rivals the state-of-the-art methods. In addition, our method can complete the task with less dialogue turns and achieves a better distinguishing capability on diseases with similar symptoms.

\end{itemize}

\begin{figure*}[!htbp]
\centering
\includegraphics[width=2\columnwidth]{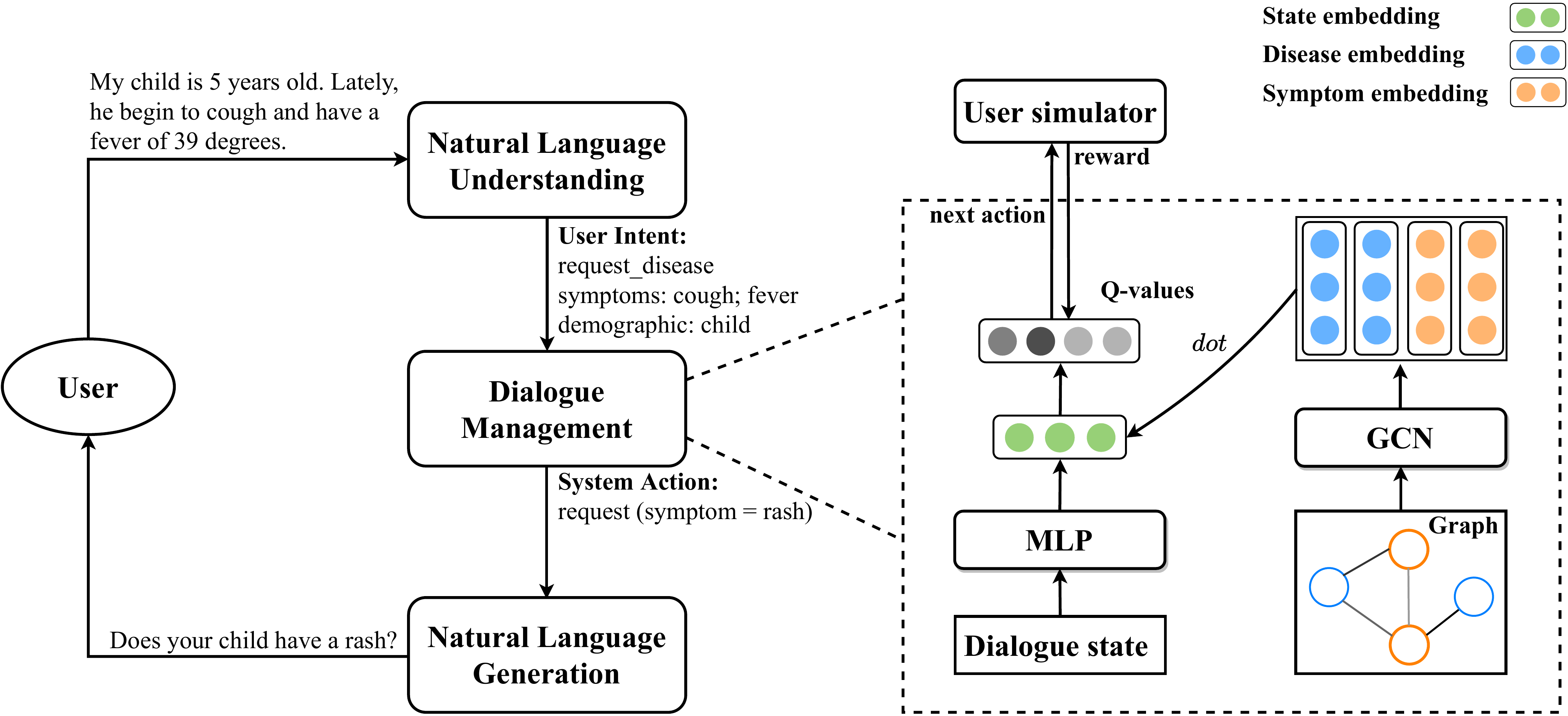} 
\caption{Illustration of the proposed graph based dialogue system. The left part of this figure shows the overview of the proposed dialogue system, and the right part is the details of Graph-DQN for DM.}
\label{model}
\end{figure*}

\section{Related Work}
Recently, knowledge based dialogue systems have attracted increasing research interest. Eric \textit{et al.}\cite{EricKCM17} regarded the smooth docking between the system and the knowledge base as a key issue and they put forward an end-to-end model with a key-value retrieval mechanism. Madotto \textit{et al.}\cite{FungWM18} raised the memory-to-sequence (Mem2Seq) that introduces the multi-hop attention mechanism to copy words directly from dialogue history or knowledge base. Many works also achieve promising performance by incorporating knowledge graph into dialogue systems\cite{ZhouYHZXZ18, MoonSKS19}. Knowledge graph stores knowledge in the form of triples that could represent the type and direction of relation between entities. However, this way of knowledge representation may be not optimal for disease diagnosis tasks.

Another area related to this work is representation learning. Representation learning aims to learn expressive latent representations from raw data for downstream tasks \cite{bengio2013representation, ChenTRY14, JiangLCC18}. Since graph-structured data allows the relational knowledge of entities to be stored and accessed efficiently, graph representation learning has led to new state-of-the-art results in many domains, including chemical synthesis, dialogue systems, recommender systems and so on\cite{tu2018unified, YingHCEHL18,MoonSKS19, HaoLHWHLCL20, yang2020learning}. Considering the powerful expression ability of graph structure, we describe the medical knowledge in the form graph, and introduce the GCN to make the constructed weighted heterogeneous graph collaborate with the Deep Q-network (DQN) \cite{mnih2015human}. 

In addition, three works are much related to this paper, in which Deep Q-network (DQN) is applied for automatic diagnosis. Kao \textit{et al.}\cite{KaoTC18} regarded the symptom checking as sequential decision problem and utilized hierarchical DQN to solve it. Wei \textit{et al.}\cite{WeiLPTCHWD18} proposed a disease diagnosis dialogue dataset and applied the DQN for disease diagnosis. The two works are mainly based on data-driven methods. Wu \textit{et al.}\cite{WuMHXSF19} raised a Knowledge-routed Deep Q-network (KR-DQN) with constraints for diseases and symptoms relation which can reduce the inquiries of illogical or repeated symptoms. But KR-DQN highly rely on statistical features and may be still insufficient in the utilization of medical knowledge. Unlike the above models, we construct a weight heterogeneous graph based on clinical dialogues to describe the disease-symptom relations and symptom-symptom relations, and propose Graph-DQN that can learn the embeddings of symptoms and diseases in the constructed graph.

\section{Method}
\subsection{Overview}
The overview of the proposed weighted heterogeneous graph based dialogue system is illustrated in Figure \ref{model}. The proposed dialogue system consists of three components: natural language understanding component (NLU), dialogue manager (DM) and natural language generation component (NLG). Given an utterance from a user, NLU detects the user intent and fills slots to represent the user intent. DM is the brain of a dialogue system, it decides which symptom to inquiry or makes a diagnosis. NLG converts system actions into natural language. The focus of this work is the DM. We construct a a weighted heterogeneous graph and present a graph based deep Q-network (Graph-DQN) for DM. NLU and NLG components are implemented with template-based methods. 
The following subsections will describe the construction of the weighted heterogeneous graph and the Graph-DQN for DM in detail.
\subsection{Graph Construction}
Given a disease diagnosis dialogue corpus, diseases and symptoms occur in the corpus are taken as nodes of a graph. There are two types of edges in the constructed graph. One is between a disease node and a symptom node, the other is between a symptom node and a disease node. To utilize global symptom co-occurrence information, this work employs point-wise mutual information (PMI) to calculate the edge weight between two symptom nodes. \begin{equation}\label{...}
PMI(i,j)=log\frac{p(i,j)}{p(i)p(j)}
\end{equation}
\begin{equation}\label{...}
p(i,j)=\frac{\# C(i,j)}{\# C}
\end{equation}
\begin{equation}\label{...}
p(i)=\frac{\# C(i)}{\# C}
\end{equation}
where $\# C$ is the total number of dialogues in a corpus, $\# C(i,j)$ is the number of dialogues that contain both symptom $i$ and $j$ in the corpus, $\# C(i)$ is the number of dialogues that contain symptom $i$. A higher PMI value implies a higher correlation between two symptoms, while a negative PMI value indicates there is little or no correlation between two symptoms.

Inspired by the term frequency–inverse document frequency (tf-idf) widely used in information retrival systems, this work presents a symptom weighting schemes, symptom frequency-inverse disease frequency (sf-idf) to calculate the edge weight between a symptom node and a disease node. The symptom frequency (sf) is proportional to the number of times the symptom appears in the disease, the inverse disease frequency (idf) is the log of the inverse probability of diseases that contain the symptom.
\begin{equation}\label{...}
sf-idf(i,j)=sf(i,j)*idf(i,j)
\end{equation}
\begin{equation}\label{...}
sf(i,j)=\frac{n_{ij}}{\sum_{k}n_{kj}}
\end{equation}
\begin{equation}\label{...}
idf(i)=log\frac{|D|}{|{j:s_i\in d_j}|}
\end{equation}
where $n_{ij}$ is the number of times that symptom $i$ occur in the disease $j$, the denominator is the total number of symptoms in the disease $j$. $|D|$ is the number of types of diseases occur in a corpus, $|{j:s_i\in d_j}|$ is the number of diseases with symptom $i$. For example, if a corpus has 4 types diseases, and a symptom $i$ occurs in 2 of them, then $|D|$ is 4 and $|{j:s_i\in d_j}|$ is 2. 
Formally, the edge weight between two nodes is defined as:
\begin{equation}\label{...}
A_{ij} =
            \begin{cases}
                PMI(i,j),  & \text{$i,j$ are symptoms, $PMI(i, j) > 0$} \\
                sf-idf(i,j), & \text{$i$ is symptoms, $j$ is disease} \\
                sf-idf(j,i), & \text{$j$ is symptoms, $i$ is disease} \\
                1, & i=j \\
                0, & \text{otherwise}
            \end{cases}
\end{equation}

In this way, we make connections of diseases and symptoms through graph paths instead of viewing them as isolated, and the strong expressive power of graph can be explored. 
\subsection{Graph-DQN for Dialogue Management}

The decision making processes in the DM is often cast as a Markov Decision Process (MDP) \cite{young2013pomdp} and can be trained via reinforcement learning (RL). Deep Q-network (DQN) \cite{mnih2015human} is a classical RL methods, which combines RL with deep learning. This subsection introduces a novel Graph-DQN that can model the graph data to obtain more reasonable actions (see the right part in Figure \ref{model} for an illustration). 

In this work, the dialogue system action space size $n = num\_greeting+M+N$, where $num\_greeting$ denotes the number of greeting action, $M$ is the number of optional diseases and $N$ is the number of optional symptoms. The Graph-DQN takes dialogue state $s$ and the weighted heterogeneous graph $G$ as inputs and outputs the $a_r \in \Bbb R^n$ that contains the Q-values (also known as action-value) of optional actions:
\begin{equation}
a_r= Q (s, G|\theta),
\end{equation}
where $\theta$ is the parameter of Graph-DQN. The Q-value of an action means the expected accumulated reward from that state. The dialogue state $s$ contains the one-hot representation of the previous action of both dialogue system and user, known symptoms and the current turn count. Multi-Layer Perceptron (MLP) is employed to obtain a hidden representation $s_h$ by taking the concatenation of the representations of state $s$ as input. The structure of the MLP is a neural network with two layers.

Motivated by the convincing performance of the Graph Convolutional Network (GCN) on the graph data \cite{HamaguchiOSM17, HamiltonYL17}, we introduce it to learn the embeddings of diseases and symptoms in the weighted heterogeneous graph. Formally, let $G = (V, E)$ denote the weighted heterogeneous graph, with $V (|V| = n, \text {the greeting actions serve as separate nodes}.)$ and $E$ being the sets of nodes and edges, respectively. $X \in \Bbb R^{n\times d}$ is the feature matrix, each row $x_v \in \Bbb R^d$ is the feature vector for the node $v$, where $d$ is the dimension of feature vectors. $X$ is simply set as an identity matrix which means every disease or symptom is represented by a one-hot vector. Let $A \in \Bbb R^{n\times n}$ be the adjacency matrix of $G$, where $A_{ij}$ equals the edge weight of the edge going from node $i$ to node $j$. If there is no edge between node $i$ and node $j$, then $A_{ij}$ is zero. $D \in \Bbb R^{n\times n}$ is used to denote the degree matrix of $G$, where $D_{ii} = \sum_j A_{ij}$. This work utilizes a one-layer GCN, the $k$-dimensional node feature is computed as
\begin{equation}
H=\sigma(\tilde{A} X W),
\end{equation}
\begin{equation}
\tilde{A}=D^{-\frac{1}{2}}AD^{-\frac{1}{2}}
\end{equation}
where $W \in\Bbb R^{d\times k}$ is trainable weight matrices and $\sigma$ is a nonlinear function (e.g., ReLU). $H \in \Bbb R^{n\times k}$ consists of embeddings of $n$ optional actions.

Then we obtain Q-values of candidate actions by computing the inner product between $H$ and $s_h$:
\begin{equation}
a_r = H \cdot s_h^\top.
\end{equation}
Finally, the DM will select the action with the largest Q-value as the next action.

Following \cite{mnih2015human}, two important tricks for training DQN, target network and experience replay are applied in this work. The DM's experience $e_t (s_t , a_t , r_t , s_{t+1})$ at each time-step $t$ is stored in a experience replay buffer. Let $y$ denote the expected Q-value of taking an action $a$ under state $s$, according to the Bellman equation \cite{bellman2015applied}:
\begin{equation}
y=r+\gamma max_{a^{\prime}}Q(s^{\prime},a^{\prime}|\theta_{target}),
\label{b}
\end{equation}
where $r$ denotes the reward\footnote{In DQN framework, the agent will obtains an immediate reward $r \in \Bbb R$ after taking an action $a$. The setting of reward value is introduced in the Deployment Details section.} after taking action $a$, the state $s$ transfers to $s^{\prime}$ after taking the action $a$. $\theta^{\prime}$ denotes the parameters of the target network, which are updated merely every $C \in \Bbb N$ iterations with the assignment $\theta^{\prime} = \theta$ and $\gamma$ is a discount rate. This work uses the Huber loss function \cite{huber1964robust}:
\begin{equation}
L = \begin{cases} \frac{1}{2}(y-Q (s, a|\theta))^2, & |y-Q (s, a|\theta)| \leq \alpha\\ \alpha(|y-Q (s, a|\theta)|-\frac{1}{2}\alpha), & |y-Q (s, a|\theta)| > \alpha \end{cases},
\label{loss}
\end{equation}
where $\alpha \in \Bbb R^{+}$ controls the transition between two functions. Besides, we use $\epsilon$-greedy exploration policy in training phase. It chooses an action randomly with probability $\epsilon$ and takes the action given by $argmax_a Q(s, a|\theta)$ with probability $1 - \epsilon$.

\section{Empirical study}
\subsection{Dataset}
\subsubsection{MZ dataset} The MZ dataset \cite{WeiLPTCHWD18} was collected from the Baidu Muzhi Doctor website. This dataset contains 710 dialogue goals that involve 4 types of diseases (infantile diarrhea, children functional dyspepsia, upper respiratory infection and children's bronchitis) and 66 types of symptoms. Each dialogue goal contains the actual result of diagnosis, the symptoms in the patient’s self-report, and the symptoms obtained by conversing with the patient. The sizes of the training set and test set are 568 and 142, respectively. 
\subsubsection{DX dataset} The DX dataset \cite{XuZGLTL19} was collected from an online health-care website (dxy.com). It contains 527 real medical diagnosis dialogues. The DX dataset covers five diseases (pneumonia, allergic rhinitis, upper respiratory infection, diarrhea, and hand-foot-mouth disease) and 41 symptoms. As \cite{XuZGLTL19}, 423 dialogues are selected for training models, and the remaining 104 dialogues are regarded as the test set. 
\subsection{Experimental Configuration}
\subsubsection{Deployment Details} This work employs the proposed dialogue system with PyTorch and PyTorch Geometric \cite{abs-1903-02428}. Following \cite{WeiLPTCHWD18,XuZGLTL19}, the discount rate $\gamma$ in Equation \ref{b} is set to 0.9 and the $\epsilon$ of the $\epsilon$-greedy exploration policy is 0.1. $\alpha$ in Equation \ref{loss} is set to 5. The maximum number of dialogue turns is set to 22, the rewards for successful and failed diagnoses are set to +44 and -22, respectively. The size of experience replay buffer is set to 10000, the batch size is set to 32 and the learning rate is 0.01. The stochastic gradient descent (SGD) algorithm is used as the optimizer. The Graph-DQN is trained for 800 epochs.

A user simulator that mimics human behaviors is required to train the proposed dialogue system. The user simulator extracts a user goal from the dataset to start the diagnostic process. The user simulator has two types of action: request and notification. The request action is to request the system for diagnosis. The notification action is to answer the inquiry of a certain symptom, which consists of three answers, True, False, and Not sure. The dialogue will be terminated and judged as a successful diagnosis if the dialogue system informs a correct disease, and when the system informs a wrong disease or the current dialogue reaches the maximum number of turns, it will be judged as a failed diagnosis.     
\subsubsection{Baseline Methods}
Several methods are selected to compare with the proposed dialogue system. ``SVM-em'' means the method using Support Vector Machine (SVM) \cite{cortes1995support} to predict diseases by taking explicit symptoms as inputs.  ``SVM-em\&im'' means the SVM trained with both explicit and implicit symptoms. ``Basic DQN'' means the method employing DQN for disease diagnosis dialogue systems, which is proposed in \cite{WeiLPTCHWD18}. Sequicity \cite{lei2018sequicity} is an outstanding end-to-end task-oriented dialogue system framework. ``DQN+KB+ RB'' \cite{XuZGLTL19} has a knowledge branch and a relation branch, and the two branch are combined them with DQN. ``PG+MI+GAN'' \cite{XiaZSLH20} is a policy gradient framework based on the Generative Adversarial Network, and adds mutual information to enhance the reward function. Furthermore, Graph-DQN that uses an unweighted heterogeneous graph is referred as ``Graph-DQN (unweighted graph)''
.  ``SVM-em'' and ``SVM-em\&im'' are only trained for MZ dataset and Sequicity is only trained for DX dataset. This is because MZ dataset only contains user-goals and thus can not be trained with an end-to-end dialogue system; DX dataset that reserves original conversation data can train an end-to-end dialogue system.

\subsection{Results and Discussion} 

\begin{table}[!htbp]
\centering
\caption{Performance comparison on the MZ dataset.}
\label{mz}
\begin{tabular}{ccc}
\toprule
Method & Accuracy &Ave turns \\
\midrule
SVM-ex & 0.59 &- \\
SVM-ex\&im & 0.71 &-\\
Basic DQN \cite{WeiLPTCHWD18} & 0.65 & 5.11\\
DQN + KB \cite{XuZGLTL19} & 0.68 & 4.00 \\
DQN+KB+RB \cite{XuZGLTL19} & \textbf{0.73} & 3.18 \\
PG+MI+GAN \cite{XiaZSLH20} & \textbf{0.73} &-  \\
\hline
Graph-DQN (unweighted graph) & 0.676 & 2.46 \\
Graph-DQN & 0.697 & \textbf{2.24}\\
\bottomrule
\end{tabular}
\end{table}

\begin{table}[!htbp]
\centering
\caption{Performance comparison on the DX dataset.}
\label{dx}
\begin{tabular}{ccc}
\toprule
Method & Accuracy & Ave turns \\
\midrule
Sequicity \cite{lei2018sequicity} &0.285 & 3.40\\
Basic DQN \cite{WeiLPTCHWD18} & 0.731 & 3.92 \\
DQN + KB\cite{XuZGLTL19} & 0.740 & 3.56\\
DQN+KB+RB \cite{XuZGLTL19} & 0.740 & 3.36 \\
PG+MI+GAN \cite{XiaZSLH20} & \textbf{0.769} & 2.68 \\
\hline
Graph-DQN (unweighted graph) & 0.731 & 2.52 \\
Graph-DQN & 0.740 &\textbf{2.21}\\
\bottomrule
\end{tabular}
\end{table}

\begin{figure}[!htbp]
\centering
\includegraphics[width=1\columnwidth]{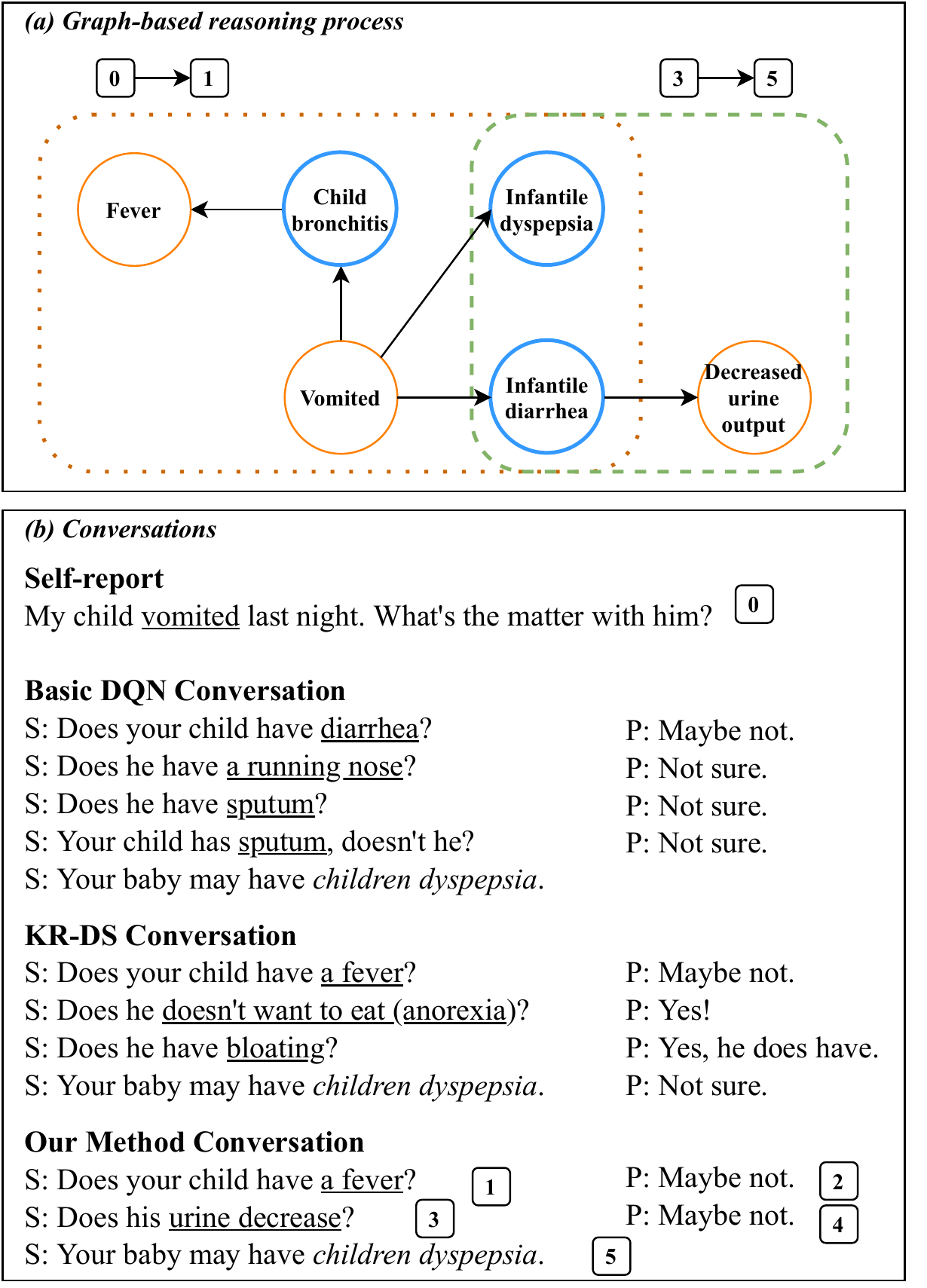} 
\caption{(a) depicts the graph-based reasoning process. (b) shows conversations from different methods. Conversations from our method are annotated and grounded with a part in (a). Specifically, except dyspepsia and diarrhea, bronchitis might cause the symptom of vomiting. Therefore, our method first rules out the possibility of bronchitis by asking if the patient has a fever (left part in (a)). Then, to discriminate between dyspepsia and diarrhea, our method inquiries the urine volume of the patient since decreased urine volume is not a typical symptom of dyspepsia but a typical symptom of diarrhea (right part in (a)).}
\label{example}
\end{figure}

The performance comparisons of different methods on MZ and DX datasets are shown in Tables \ref{mz} and \ref{dx}, respectively. From Tables \ref{mz} and \ref{dx}, we have the following findings.

First, the effectiveness of the proposed Graph-DQN. Graph-DQN outperforms DQN in both datasets, and it is worthy of note that our method beats DQN by more than 5\% in MZ dataset. The performance improvement for DX dataset is comparatively slight. This may because the size of DX dataset is relatively small, which may restrict the performance of GCN. In addition, Graph-DQN outperforms the “DQN + knowledge branch", which implies that the features about medical knowledge learnt by Graph-DQN are richer than the statistical features.  

Second, our method completes the task with less turns. In fact, baseline methods tend to ask symptoms having low degree of distinction, and thus increase dialogue turns. Figure \ref{example} illustrates examples from the basic DQN, KR-DS and our method. One can observe that DQN asks some irrelevant symptoms. KR-DS performers better but the symptoms it asked have poor discrimination. Our method can inquiries the discriminative symptoms for diseases with similar symptoms and completes the task with less turns. This phenomenon demonstrates that our method could capture symptom-disease
relations and symptom-symptom relations better. We attribute this advantage to the utilization of the weighted heterogeneous graph. 

Third, the effectiveness of the devised edge weights of the constructed graph. We can find that using the unweighted graph decreases the diagnostic accuracy and increases the dialogue turns consistently. This indicates that PMI and the presented sf-idf may provide more information for disease diagnoses.

\subsection{Qualitative Analysis}

\begin{figure*}[!htbp] 
	\centering  
	\vspace{-0.35cm} 
	\subfigtopskip=2pt 
	\subfigbottomskip=10pt 
	\subfigcapskip=-5pt 
	\subfigure[Result of KR-DS on the MZ dataset]{
		\label{krds-mz}
		\includegraphics[width=0.4\linewidth]{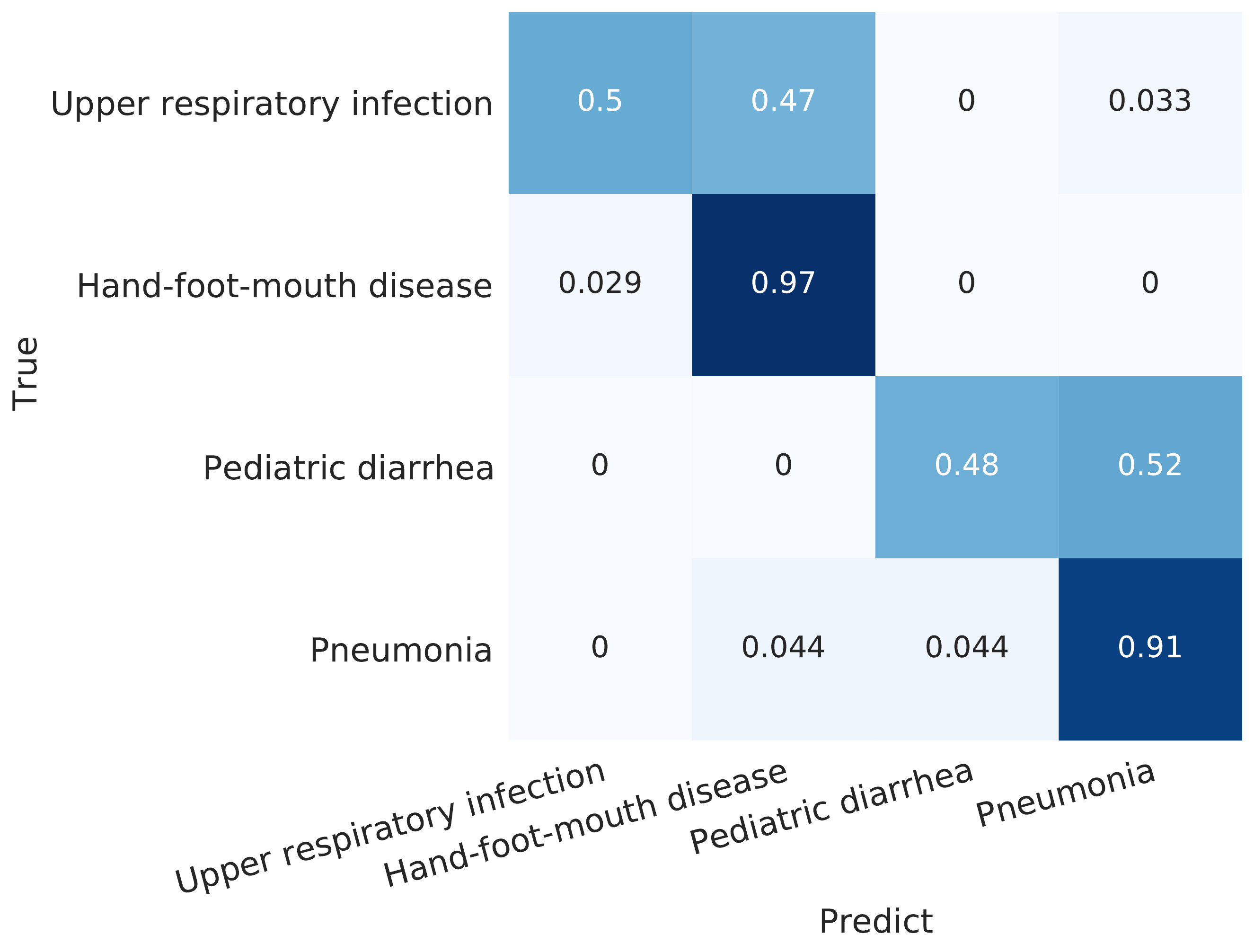}}
	\quad 
	\subfigure[Result of Graph-DQN on the MZ dataset]{
		\label{krds-mz}
		\includegraphics[width=0.4\linewidth]{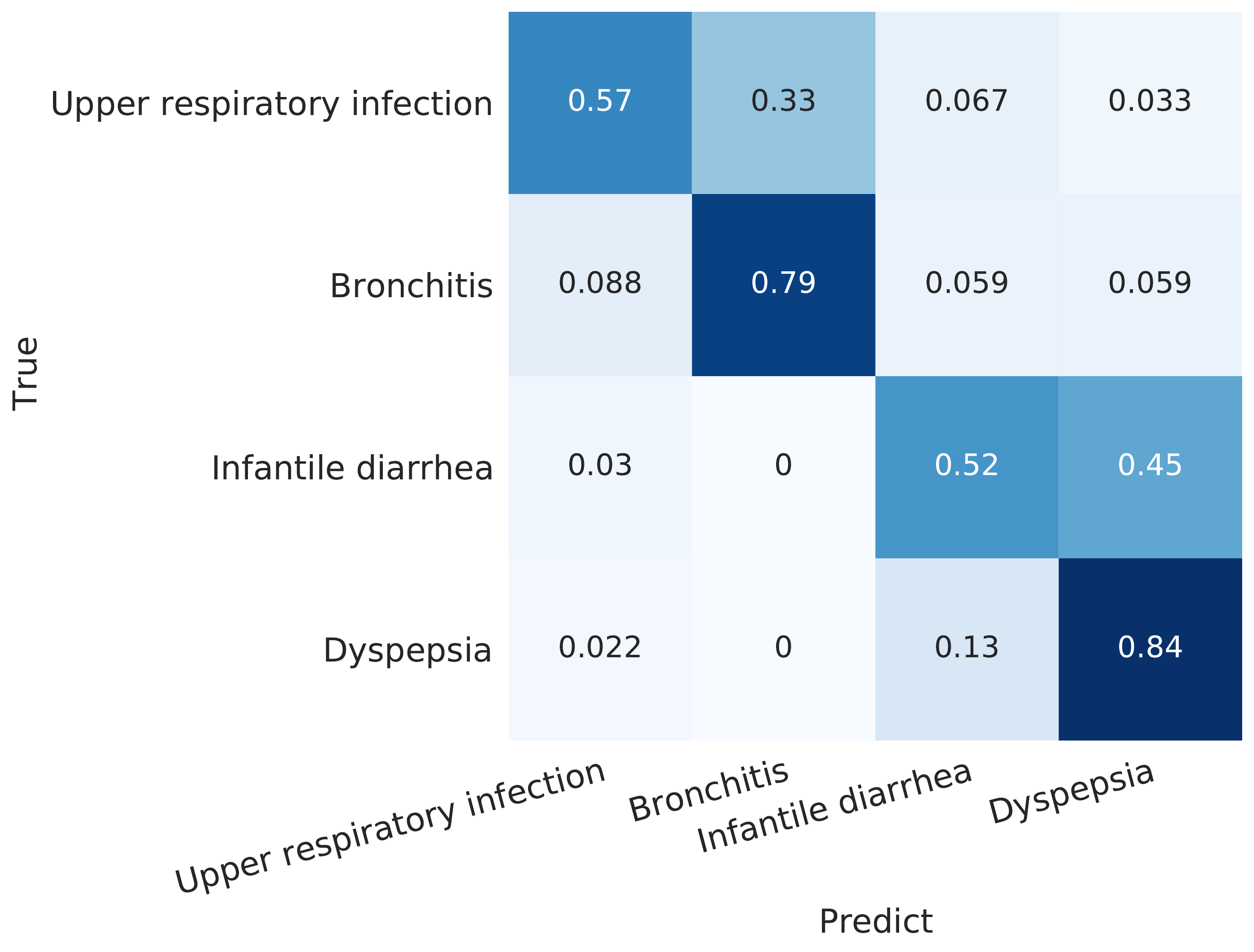}}
	\subfigure[Result of KR-DS on the DX dataset]{
		\label{level.sub.3}
		\includegraphics[width=0.4\linewidth]{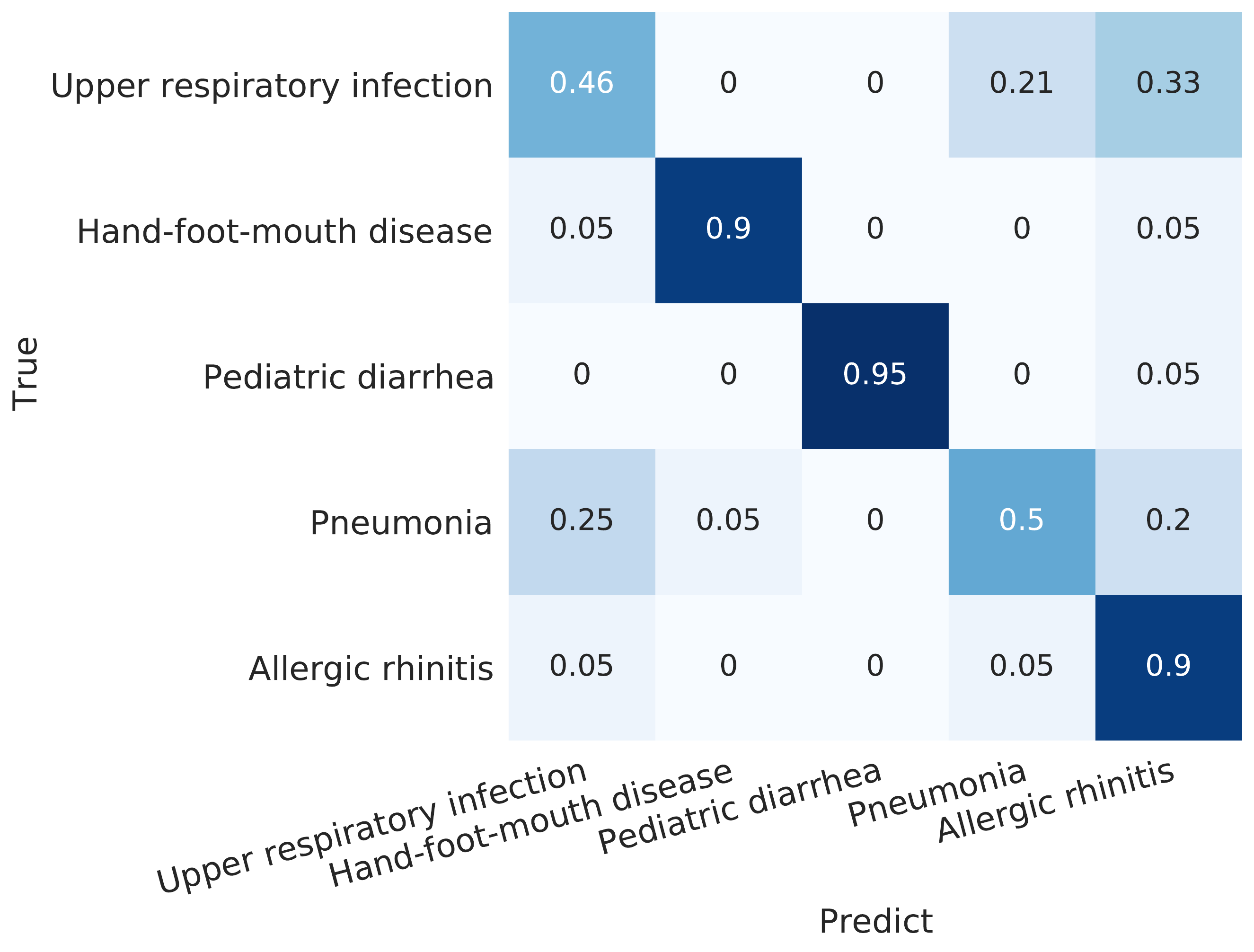}}
	\quad
	\subfigure[Result of Graph-DQN on the DX dataset]{
		\label{}
		\includegraphics[width=0.4\linewidth]{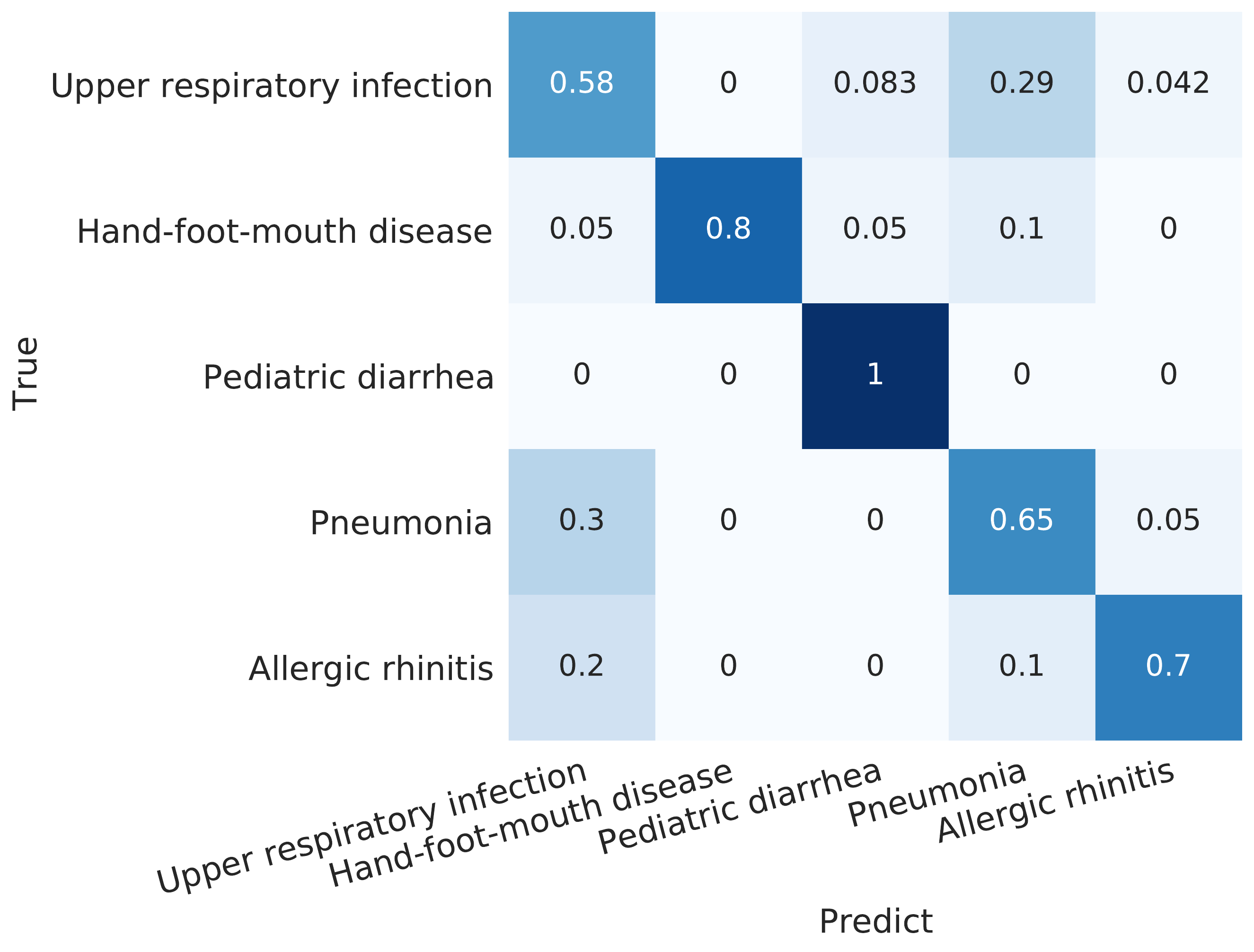}}
	\caption{Normalized confusion matrices for the results of KR-DS and Graph-DQN}
	\label{cm}
\end{figure*}

Figure \ref{cm} visualizes detailed results of KR-DS and our method by confusion matrices. From these figures, one can see that KR-DS is difficult to distinguish diseases with similar symptoms. For example, KR-DS misdiagnose many upper respiratory infection cases as bronchitis or allergic rhinitis. However, Graph-DQN could better distinguish between diseases with similar symptoms. Therefore, the proposed method may have a higher clinical value. 
\begin{figure*}[!htbp] 
	\centering  
	\vspace{-0.35cm} 
	\subfigtopskip=2pt 
	\subfigbottomskip=10pt 
	\subfigcapskip=-5pt 
	\subfigure[DQN trained on the MZ dataset]{
		\label{}
		\includegraphics[width=0.4\linewidth]{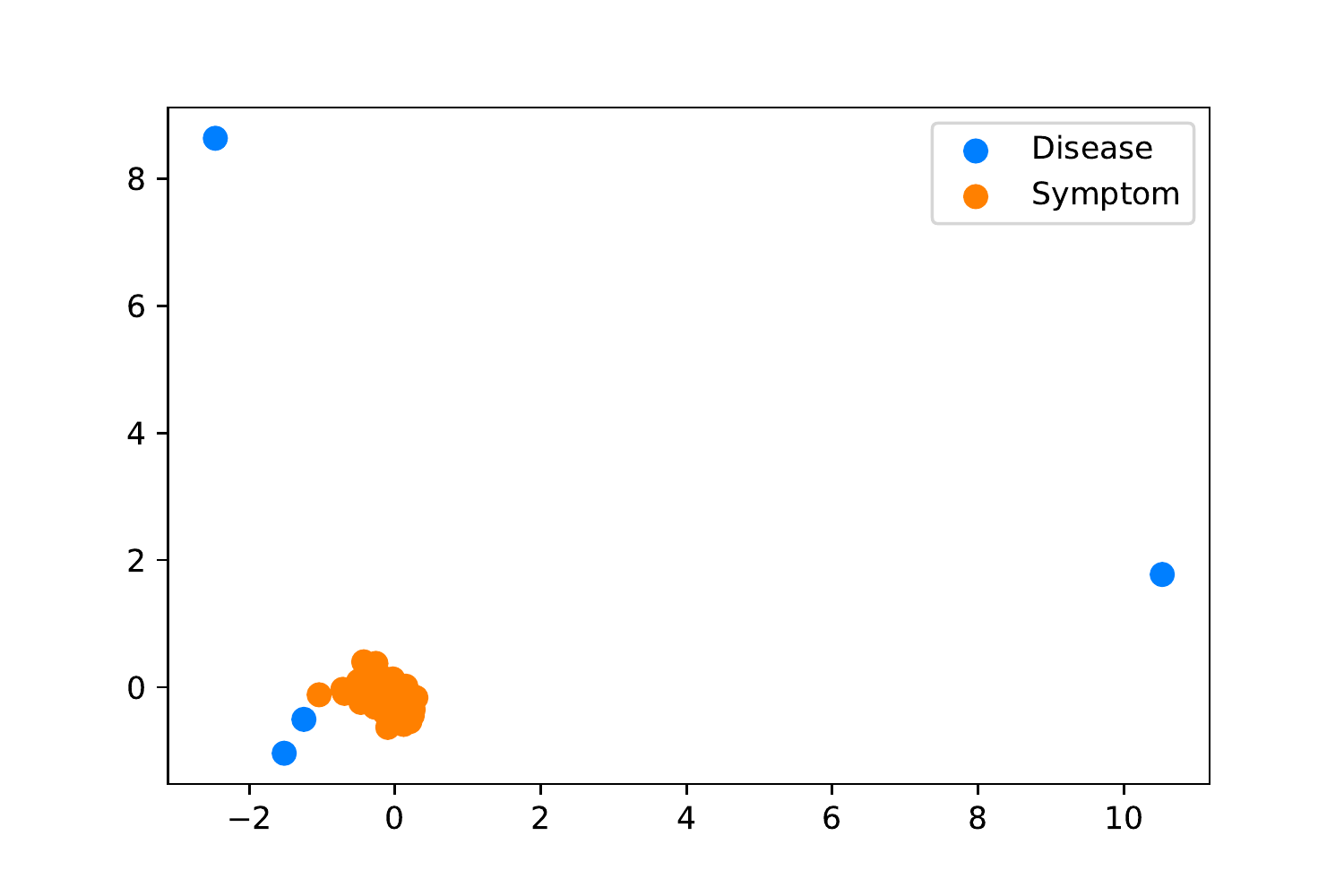}}
	\quad 
	\subfigure[Graph-DQN trained on the MZ dataset]{
		\label{}
		\includegraphics[width=0.4\linewidth]{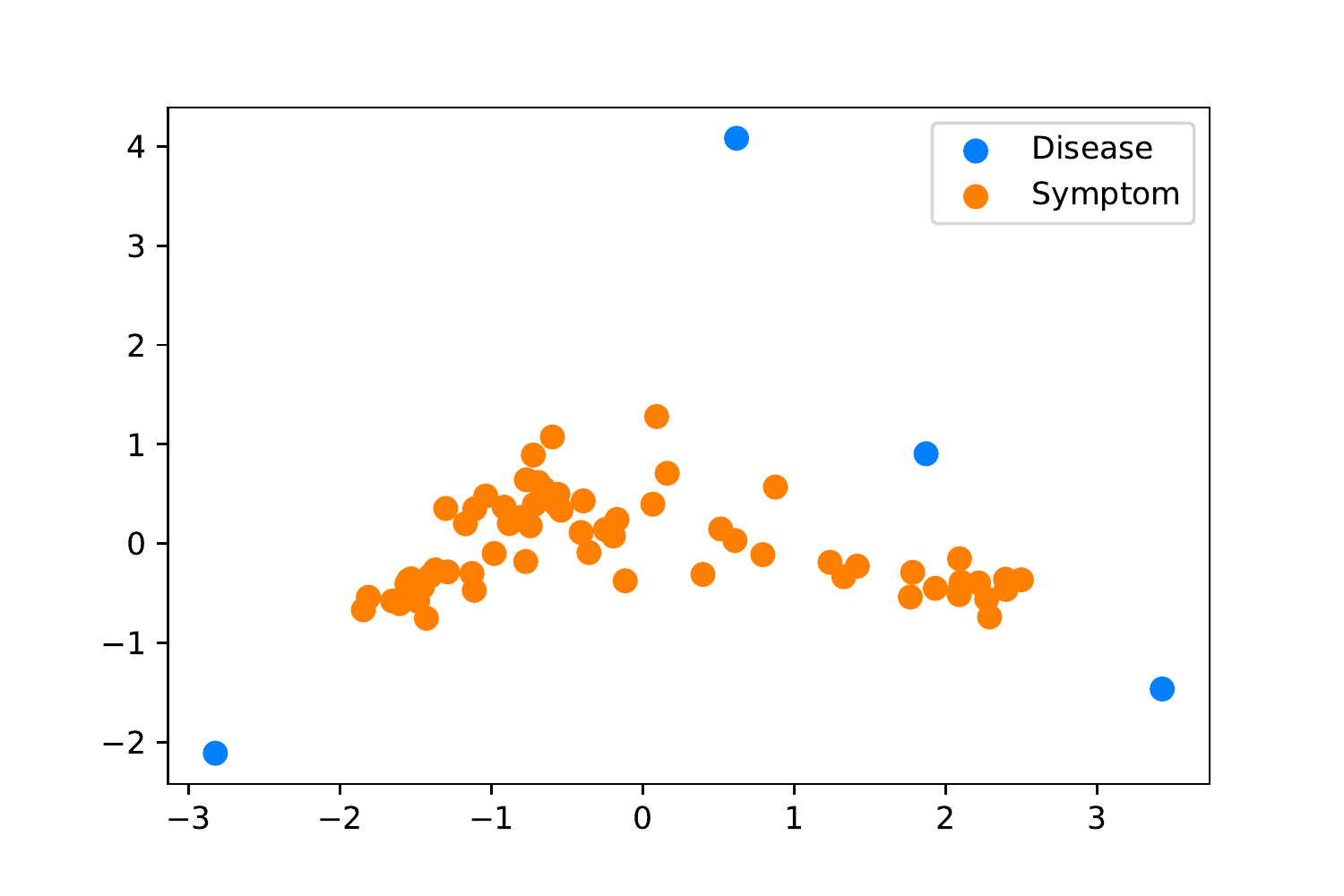}}
	\subfigure[DQN trained on the DX dataset]{
		\label{level.sub.3}
		\includegraphics[width=0.4\linewidth]{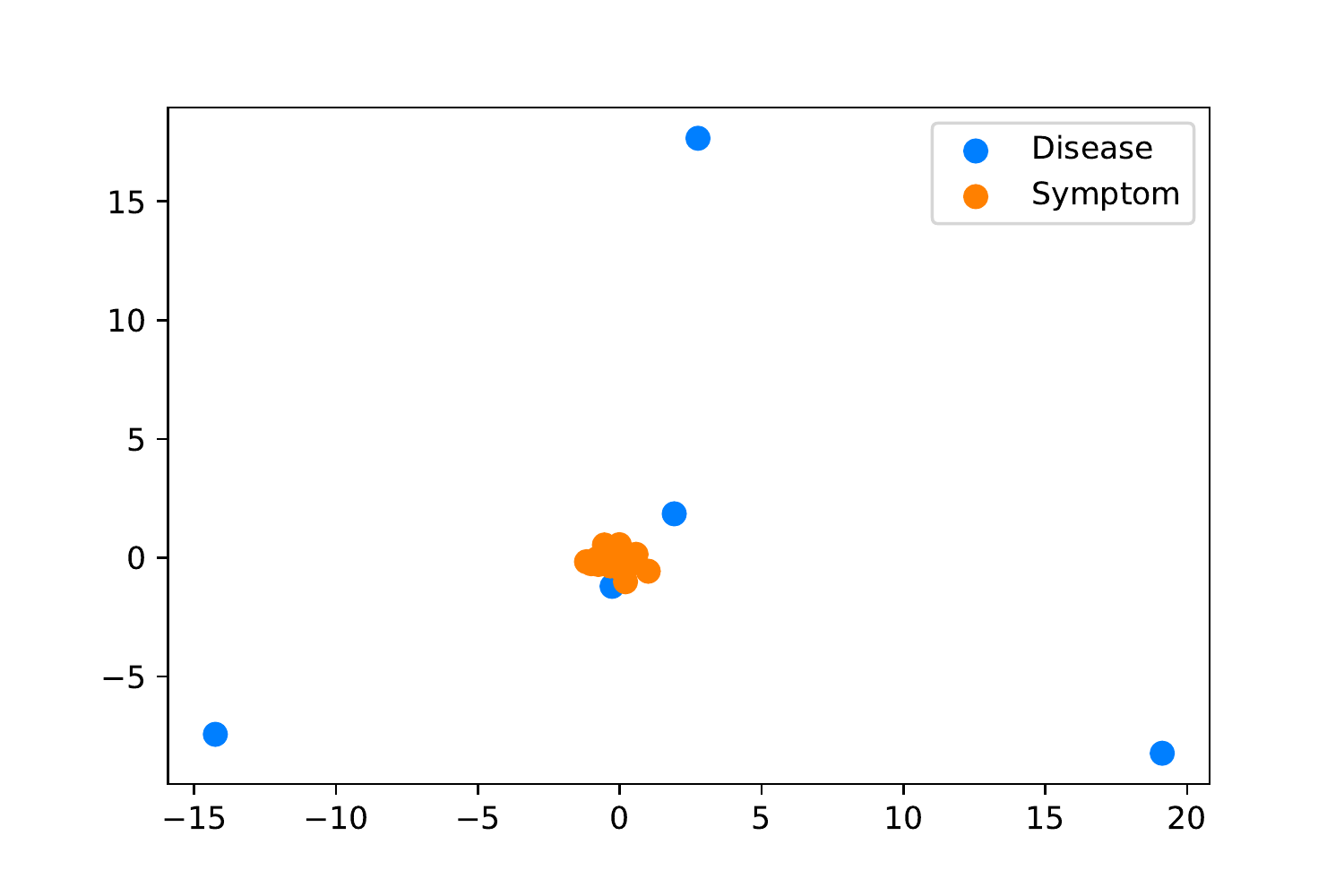}}
	\quad
	\subfigure[Graph-DQN on the DX dataset]{
		\label{}
		\includegraphics[width=0.4\linewidth]{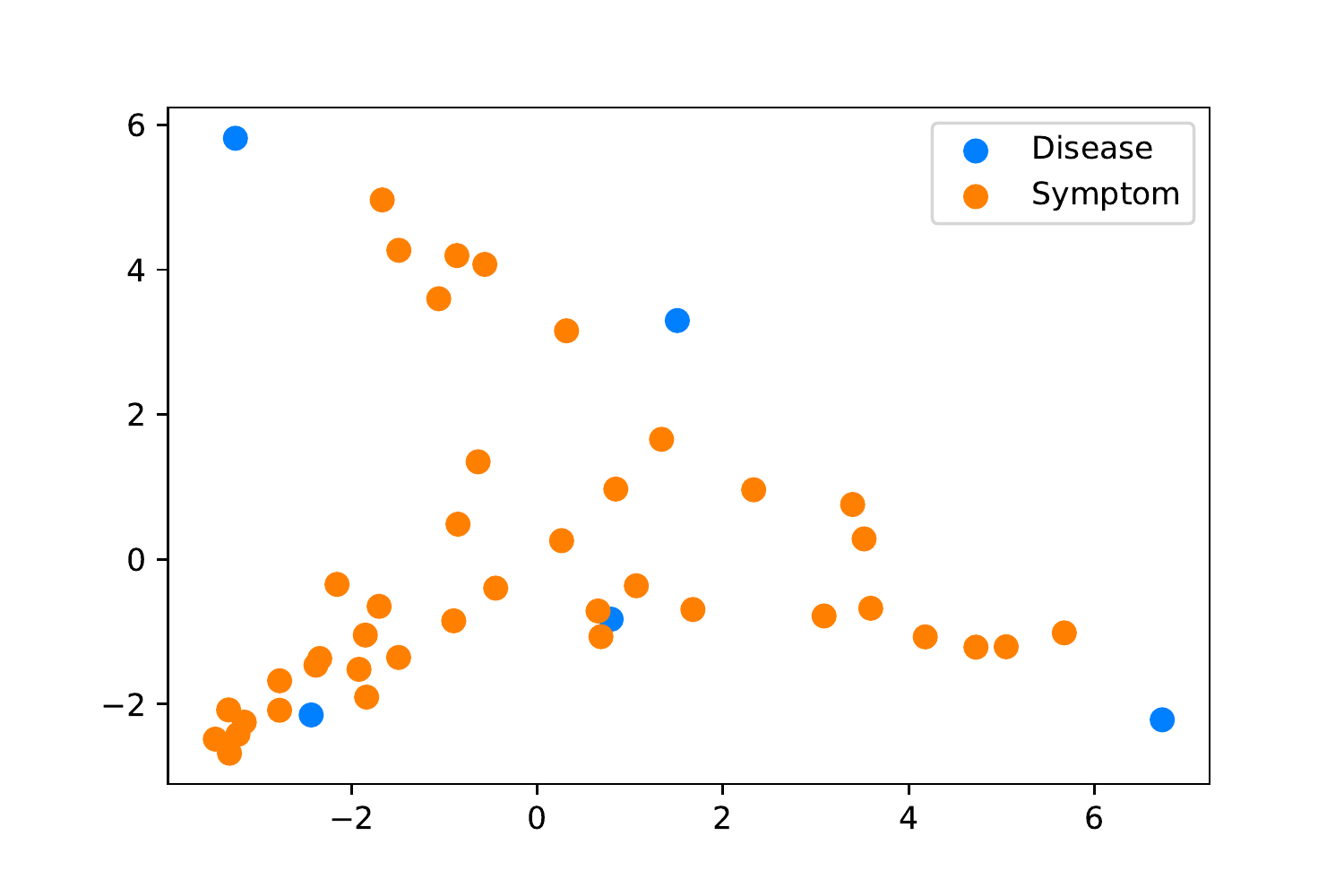}}
	\caption{Visualization of Symptom Embeddings and Disease Embeddings Learnt by DQN and Graph-DQN}
	\label{embedding}
\end{figure*}

To understand Graph-DQN and DQN intuitively, we visualize the embeddings of diseases and symptoms learnt by Graph-DQN ana DQN. To be specific, we utilize PCA to layout a embedding vector in a 2-dimensional space, and report the results in Figure \ref{embedding}. As we can see, the diseases and symptoms embeddings learnt by Graph-DQN gather together into clusters. Moreover, the embedding of a disease is close to the embeddings of its common symptoms. In contrast, the embeddings learnt by DQN are hardly split into different groups.

\section{Conclusion}
This paper presents a weighted heterogeneous graph based dialogue system for disease diagnosis. Instead of relying on statistical features, we devise a weighted heterogeneous graph to describe the symptom-disease and symptom-symptom relations, and propose a graph based deep Q-network (Graph-DQN) for dialogue management. By integrating the constructed weighted heterogeneous graph, Graph-DQN could learn symptom co-occurrence information and the importance of a symptom to the diagnosis of a disease. Experiments conducted on two datasets confirm the effectiveness of the proposed dialogue system by utilizing the constructed weighted heterogeneous graph. Besides, experiments demonstrate that the proposed dialogue system can complete the diagnosis task with less dialogue turns and achieve better distinguishing capability on diseases with similar symptoms, which means our method may be more competitive in clinical applications. In the future, we would like to further explore the potential of GNNs in the research of medical diagnosis. We would also like to incorporate information of patients into disease diagnosis agents, such as the heredity and demographics of patients.


\bibliographystyle{IEEEtran}
\bibliography{IEEEabrv,bibtex.bib}
\end{document}